\documentclass{ecai} 

\usepackage{latexsym}
\usepackage{amssymb}
\usepackage{amsmath}
\usepackage{amsthm}
\usepackage{booktabs}
\usepackage{enumitem}
\usepackage{graphicx}
\usepackage{color}

\newcommand{\BibTeX}{B\kern-.05em{\sc i\kern-.025em b}\kern-.08em\TeX}

\begin{document}


\begin{frontmatter}


\paperid{1794} 


\title{DeepClean: Integrated Distortion Identification and Algorithm Selection for Rectifying Image Corruptions}


\author[A]{\fnms{Aditya}~\snm{Kapoor}\thanks{Corresponding Author Email: aditya.kapoor1@tcs.com.}}
\author[A]{\fnms{Harshad}~\snm{Khadilkar}}
\author[A]{\fnms{Jayvardhana}~\snm{Gubbi}} 

\address[A]{Tata Consultancy Services}


\begin{abstract}
Distortion identification and rectification in images and videos is vital for achieving good performance in downstream vision applications. Instead of relying on fixed trial-and-error based image processing pipelines, we propose a two-level sequential planning approach for automated image distortion classification and rectification. At the higher level it detects the class of corruptions present in the input image, if any. The lower level selects a specific algorithm to be applied, from a set of externally provided candidate algorithms. The entire two-level setup runs in the form of a single forward pass during inference and it is to be queried iteratively until the retrieval of the original image. We demonstrate improvements compared to three baselines on the object detection task on COCO image dataset with rich set of distortions. The advantage of our approach is its dynamic reconfiguration, conditioned on the input image and generalisability to unseen candidate algorithms at inference time, since it relies only on the comparison of their output of the image embeddings.
\end{abstract}

\end{frontmatter}


\section{Introduction}

Due to recent advances in Machine Learning, images and videos have been extensively used in the applications of robotics, healthcare, satellite imagery, social media, and security to name a few. In each of these applications, the quality of image and video plays a vital role in achieving good performance in downstream vision tasks like image classification, object detection and segmentation. Even before the images are used for a particular task, they need to be acquired, processed, compressed, transmitted, stored and then retrieved. The quality of the image may deteriorate in either of the aforementioned stages and hence can introduce different kinds of distortions that can affect the downstream vision task's performance. For example, during acquisition the image sensor or circuitry of a digital camera or scanner can inject noise in the image. If an image is being taken with a hand-camera, motion blur can be introduced in the image. Some image compression algorithms are lossy and may erase information that maybe crucial.  While transmission, data packets can be lost which might degrade the image quality. As a result, identification and rectification of the distortions in the image is very useful to ensure decent performance of the end vision goal.

Currently, to the best of our knowledge, in the community most experts still configure the image processing pipeline by visually inspecting the image for every specific use case. If we were to break down the expert methodology into a procedure, at first the expert identifies the possible distortions that are present in the image and then later uses existing algorithms in their toolkit to rectify the distortions in the image with trial and error. Once the experts arrive at a pipeline, it remains fixed for the lifetime of the task. Since this process is manual and heavily relies on intuition, we believe there is a need to automate this process since the image quality heavily impacts the downstream task. If the image processing pipeline is not carefully configured, it can cause more harm than good. In Sec \ref{Experiments} we show empirically how a wrong image processing pipeline can heavily impact the metrics of object detection. It is also not straightforward to determine the choice of algorithms to be used to recover the image. For example, to restore an image if it needs to undergo exposure correction and then denoising, the choice of algorithms also plays a vital role. Two individually best performing algorithms might not yield the best result when used in conjunction with each other. As a result algorithm choice also becomes crucial alongside distortion identification. Finally, most image processing pipelines are fixed for the entire dataset which is not ideal since different images in the same dataset might require different sequence and choice of correction algorithms for rectification.

As a result, this paper attempts to deal with distortion identification and correction algorithm selection for every input image. We believe that such a framework can be very useful to the computer vision community.

Our proposed framework has the following main contributions:
\begin{itemize}
    \item We propose DeepClean, a plug-and-play data-driven simple multi-task framework that automatically identifies distortions and dynamically selects appropriate algorithms in the vision pipeline without any manual engineering efforts.
    \item Our framework does not assume any prior knowledge about the pool of candidate algorithms to choose from and hence has generalization capabilities.
    \item We also experiment with new forms of distortions that the framework has not seen during training time and show that it adapts well to such situations.
    \item We demonstrate the advantage of DeepClean compared to baselines on the COCO dataset on object detection as the downstream vision task.
\end{itemize}


\section{Related Work}

In this section, we review some prior works that attempt to identify image distortions. Some of these works consider distortion identification as the primary task while some consider it to be a secondary task of an image quality assessment (IQA) process. Further, we also explore other algorithm selection frameworks that have been used to identify the best single-agent and multi-agent path finding algorithm given a particular problem instance described using an image. Lastly, we also look into AI-planning and Reinforcement Learning (RL) methods that have been previously used to address a similar problem setting.

\subsection{Distortion Classification}


Namhyuk Ahn et al. \cite{namhyuk2018} used CNN architectures pretrained on ImageNet dataset \cite{jia2009imagenet} to classify distortion type present in the input image without a clean reference image. Mateusz et al. \cite{Buczkowski2019} trained two CNN architectures that were similar in terms of number of layers and layout but had different size and number of parameters to predict the distortion in an image. Bianco et al. in \cite{Bianco_2021} analyzed the features extracted from different layers of different deep neural network (DNN) architectures and then evaluate their relevance using clustering. The best features were then used for the recognition of the type of distortion. Unsupervised Deep visual representations show cases the ability to recognize the various types of image distortion.

Distortion identification is useful for IQA since it can either help to select the most appropriate metric for evaluation or provide vital information about the distortions from the metric. Falk et al. \cite{Falk2007ImprovingRO} proposed to linearly combine a set of features from a pool of full-reference IQMs that were relevant for distortion identification. Moorthy et al. \cite{moorthy2010} describes a classifier that relies on the statistics of the distorted image to classify the distortion type. Peng et al. \cite{Peng2012} proposed a two step procedure for quality assessment where in the first step, a SVM was used to identify the distortion type and in the second step using the knowledge of the distortion type, three existing IQMs were fused using K-nearest neighbors. Chetouani et al. \cite{CHETOUANI2012948} proposed a framework built on the assumption that there is no single universal IQM that can effectively estimate the image quality across all different types of distortions. They used linear discriminant analysis (LDA) to classify the distortions before image quality estimation. During classification, they relied on relevant quality scores derived from different IQMs that were applied on both the clean and the distorted version of the clean image. Kang et al. \cite{Kang2015} proposed IQA-CNN, a model that estimates image quality and identifies distortion types. IQA-CNN+ \cite{Kang2015} and IQA-CNN++ \cite{Kang2015} are the two advanced versions of IQA-CNN. A CNN model was also proposed by Wang et al. \cite{Wang2016} for distortion identification and quality assessment without the clean reference image. Kede et al. \cite{Kede2018} proposed a CNN-based IQM with two sub-networks, distortion identification network and quality prediction network. Both these networks shared the early layers. At first, the distortion identification network is trained followed by quality prediction network starting from the pre-trained first network. Huang et al. \cite{Huang2019} proposed a mask gated convolutional network (MCGN) that predicts distortions as well as image quality score simultaneously. MCGN uses an encoder which is designed to capture the transformation between clean and distorted version of the clean image as low level features. Later, the high level features are extracted and utilized to predict the image quality score and distortion type. Ameur et al. \cite{Ameur2021} proposes a standalone multitask learner to identify multiple distortions present in the image. They deploy a DenseNet-169 backbone feature extractor pretrained on ImageNet dataset and subsequently learn one prediction head to predict the absence or presence of each possible distortion in the image. 

The above works achieve good results on distortion identification. However, most of the approaches rely on domain knowledge to manually design feature descriptors, almost all of them are not scalable solutions and none of them focus on selecting appropriate correction algorithms to rectify the distortion present in the image.

\subsection{Algorithm Selection in Path Finding (PF)}
Algorithm selection has been an active area of research in Single-Agent and Multi-Agent PF since different PF algorithms work well for different problem instances. Sigurdson et al. \cite{Sigurdson2017DeepLF} selects heuristic based search algorithms for single-agent path finding using a generalizable approach that makes use of image classifiers. They train AlexNet to learn a mapping between the different types of problem instances and the corresponding best path finding algorithm. Sigurdson et al. \cite{Sigurdson2019} extends his previous work and trains AlexNet to map different multi-agent path finding instances represented using RGB images to different multi-agent path finding algorithms depending on a few metrics used to measure the overall performance of the different algorithms. The authors ensure that the portfolio of the algorithms in consideration are diverse and well suited to address the different problem instances in the dataset. Similarly, Ren et al. \cite{Ren2021} also deploy a CNN with Inception module and a different problem instance definition and algorithm portfolios to address the multi-agent path finding setting. 

The above approaches try to alleviate the need for manual feature extraction and switch to CNN based approaches since they discover that CNNs are capable to learn to extract relevant features that are necessary to identify the algorithms for a given problem instance. However, the above approaches require the dataset to be re-prepared and also need to be retrained if newer algorithms are to be introduced which can be computationally expensive. On the other hand, Ren et al. \cite{Ren2021} also requires additional information alongside RGB images of the map, that is, single-agent optimal paths to improve the classification accuracy. As a result, the performance of the algorithm is conditioned on the choice of the single-agent path finding algorithm that can lead to sub-optimal predictions.

\subsection{Algorithm selection using AI planning}
The early attempts to select appropriate algorithms for automating the generation of image processing workflows using AI planning only focus on certain specific applications~\cite{chien1994using, lansky1995collage}. Nadarajan et al.~\cite{nadarajan2011flexible} proposed a system based on decomposition-based planning and ontologies for the automatic construction of video processing solutions and demonstrated it for detection and tracking in underwater videos. In this paper, the pre-processing steps only includes video capture and frame image grabbing. Similarly, Clouard et al.~\cite{clouard1999borg} proposed BORG a knowledge-rich problem-solving system that is used for dynamically constructing image processing procedures through the selection, parameter tuning and scheduling of existing algorithms. Such rule-based knowledge-driven systems can only succeed in solving problems corresponding to well-identified tasks. Sengar et al.~\cite{sengar2021automatic} developed a framework that can adapt the solutions based on changing requirements by integrating planning with RL and learning from goal-directed reward. However, the cost of each algorithm is learned based on certain image characteristics and its corresponding state abstractions rather than the entire image features directly. Similarly, Kapoor et al.~\cite{kapoor2022auto}, introduced a Transformer Architecture combined with Deep Reinforcement Learning to recommend algorithms that can be incorporated at different stages of the vision workflow. However, this approach is limited since it searches for the optimal algorithm for each preprocessing step by fixing the image processing pipeline. A recent attempt was made by Choudhury et al. ~\cite{choudhury2023methods} to couple planning and learning based approaches to construct dynamic vision pipelines for downstream tasks. 

\section{Problem Definition}
\begin{figure}[!htbp]
\centerline{\includegraphics[scale=0.4]{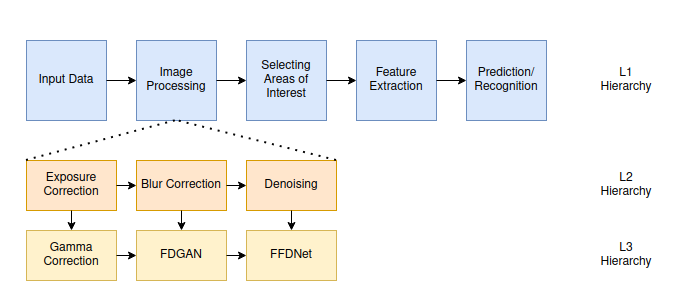}}
\caption{Description of three levels of hierarchy to generate a Vision Pipeline.}
\label{probform}
\end{figure}

We believe that there are three steps involved to compose the pipeline for a goal vision task. In Fig. \ref{probform} the first step is to define the high level pipeline which we term as \textit{L1-hierarchy}. The L1-hierarchy is a skeleton which broadly identifies the different set of sub-goals that lead to achieving the downstream vision task. In most cases, for a given goal vision task, the L1-hierarchy is task dependent and hence can be known apriori. The second step in the hierarchy, \textit{L2-hierarchy}, is to identify the arrangement of a sequence of sub-tasks that are necessary to achieve the sub-goals of the L1-hierarchy. For example, the problem of identifying the right sequence of image processing steps belongs to L2-hierarchy. Finally, we define \textit{L3-hierarchy} as the problem of identifying the algorithm to be used for each of the sub-tasks.

Let's take an example of object detection, refer Fig. \ref{probform}. Given the L1-hierarchy skeleton is fixed, the problem of L2-hierarchy is to identify the sequence of sub-tasks that need to be taken within each subgoal of the L1-hierarchy. For example, within image processing, the identification of various image processing steps like denoising, exposure-correction, blur-correction etc and their sequence. Finally, L3-hierarchy deals with the choice of the candidate algorithms that need to be employed for execution the sub-tasks like SCUNet for denoising, gamma correction for exposure rectification, etc.

The orchestration of the composition of each of the hierarchy levels has the capability to generate a vision pipeline to achieve a certain goal task. One should note that the \textit{L1-hierarchy} remains fixed (in most cases) irrespective of the nature of the input images since it's only dependent on the downstream vision task. However, the L2 and L3 hierarchy may undergo changes depending on the input images. For example, if $image_{1}$ has undergone exposure distortion and noise addition and $image_{2}$ has undergone noise addition followed by exposure distortion; the L2 and L3 hierarchy in each of the case will be different, that is denoising followed by exposure correction, and exposure correction followed by denoising respectively. Each image might also require different algorithms at each stage depending on the nature of distortion. 

In this paper, we deal with L2 and L3 hierarchy and treat them as a hierarchical sequential decision making problem. We propose a framework that can simultaneously identify the next processing step and algorithm in the sequence.

We show the effectiveness of our planning approach in Image Processing pipeline generation. However, our approach is not limited to Image Processing. Based on our definitions of the L-2 and L-3 hierarchies, the problem of identification of the correct sequence of image processing steps fall in the L2-hierarchy, and that of algorithm selection belongs to the domain of L3-hierarchy.

Mathematically, we define this problem by a tuple $\langle I, S, A, Q \rangle$, where \textit{I} $=\{i_{1}, i_{2}, .. \}$ are the set of input images, \textit{S} $=\{s_{1}, s_{2}, .. \}$ is the set of image processing steps and \textit{A} $=\{a_{1}, a_{2}, .. \}$ is the set of image processing algorithms. Further, for each image processing step in \textit{S} there is an image processing algorithm in \textit{A}. \textit{Q} $:$ \textit{S} $\times$ \textit{A} $\to \mathbb{R}$ is a function that returns the quality of the solution obtained when a image processing step $s_{i}$ $\in$ \textit{S} and a corresponding image processing algorithm $a_{i}$ $\in$ \textit{A} is chosen to process the input image $i_{i}$ $\in$ \textit{I}.

We learn a mapping function that $\pi:$ \textit{I} $\to$ $\langle S, A \rangle$ that maps each image $i_{i}\,\in$ \textit{I} to a image processing step $s_{i}\,\in$ \textit{S} in the first stage and an image processing algorithm $a_{i}\,\in$ \textit{A} in the second stage.

Our set of images \textit{I} consists of a combination of no distortions, single-level distortion or combination of two distortions. The distortions comprise of exposure changes and noise addition. Within exposure change, we either underexpose the image or overexpose the image and in case of noise addition, we add low or high levels of Gaussian noise in the images. We discuss about this in more detail in the subsequent section that describes the dataset preparation for training and testing our framework \ref{Dataset}. We fix the downstream vision task to be \textit{Object Detection} and thus fix the L1-hierarchy to be Capture RGB Image followed by Image Processing followed by Object Detection. The set \textit{S} comprises of Denoising, Underexposure Correction, Overexposure Correction and No Correction as the image processing steps and the corresponding correction algorithms are included in the set \textit{A}. Each algorithm $a_{i}$ is specialized to address a distortion of a particular type in image $i_{i}$. Thus for each image processing step $s_{i}$ $\in$ \textit{S} there exists atleast two algorithms $a_{i}$ $\in$ \textit{A}.


\section{Dataset Preparation}
\label{Dataset}
To prepare our dataset, we extend the COCO validation dataset \cite{cocodataset}. The images in our dataset fall in either of the categories; clean, distorted exposure (underexposed/overexposed), noise addition (low/high) or a combination of both distorted exposure and noise addition in different sequences. We saw that if the image were corrupted using three or more distortions, it would become really difficult to restore the image back to its original format and thus we restricted ourselves to a maximum of two levels of distortions in this study. To distort the exposure of the image, we use the Gamma correction algorithm present in Pytorch \cite{pytorch2019} which is based on the power law:
\begin{equation}
\label{eq:gamma_correction}
I_{out} = 255 \times gain \times (I_{in}/255)^{\gamma}
\end{equation}
We set the gain value to be 1 at all times and use $\gamma$ $\in$ $\{0.2, 0.8, 2.0, 3.0\}$. The former two values of gamma are used to underexpose the images and the latter two are used to overexpose the images.

To introduce noise in the images, we used Gaussian noise. The following equation was used to do the same:
\begin{equation}
\label{eq:gaussian_noise}
I_{out} = ((I_{in}/255) + \mathcal{N}(\mu,\,\sigma^{2})) \times 255
\end{equation}
We fix $\mu$ to be 0.0 throughout and sample $\sigma$ from $\{0.04, 0.08, 0.15, 0.2\}$. Again the former two values add low-level of noise in the image and the latter two values add high-level noise in the image.

Overall we generate a dataset with 202,622 images out of which 4942 are clean images, 9,884 images comprise of low and high level noise, and overexposed and underexposed images and the rest are a combination of each of the distortions in different sequences.

In case of generating the test dataset with different distortions, we adopt the following $\gamma$ values: $\{0.3, 0.9, 2.2, 3.2\}$ and $\sigma$ values: $\{0.06, 0.1, 0.17, 0.25\}$

\section{Proposed Method}
The aim of our proposed framework is to develop a robust and reliable method for distortion identification and algorithm selection for image rectification without depending on the corresponding clean images. In order to achieve this, we extend the previous work \cite{Ameur2021} which only deals with distortion identification and also make it compatible to select corresponding algorithms that can rectify the identified distortion. We first empirically show that why a deep learning approach is suitable for this method and later describe our framework in more detail.

\subsection{Motivation}
\label{Motivation}
The features obtained from a deep learning feature extractor (such as ResNet \cite{ResNet2015} or DenseNet \cite{DenseNet2016}) capture various attributes of the images in the latent space. These feature vectors can capture information about the different distortions present in the image. We empirically validate this assumption by training a ResNet-50 classifier (pretrained on ImageNet dataset) to identify the different distortion sequences present in the image. For our use-case, the classes of distortion sequence were clean image, underexposed image, overexposed image, low noise-level, high noise-level, underexposed followed by low noise-level image, low noise-level followed by underexposing the image and so on. Our classifier was able to tell apart the different distortions with 96\% accuracy. We trained both the ResNet backbone and the latter classification layer to identify the latest distortion that was added to the input image. Thus one can rely on the features extracted by the ResNet-50 feature extractor to identify different distortions present in the image.

We take inspiration from the methodology used by human experts to identify distortions in the image and select subsequent algorithms to rectify them. For the former, human experts solely rely on the visual aspect of the input image. In case of the latter, they also rely on the relative capability of the correction algorithm; that can be abstracted from metrics like PSNR \cite{PSNR} (Peak Signal-to-Noise Ratio). However, to make an algorithmic choice, human experts make a relative comparison between the different correction algorithms that are capable to rectify the existing distortion either via visually inspecting the output image of the algorithm or use some metrics well suited for the task. We use this systematic approach as the basis to construct our framework which we discuss in detail in the next subsection.

\begin{figure}[t]
\centerline{\includegraphics[scale=0.5]{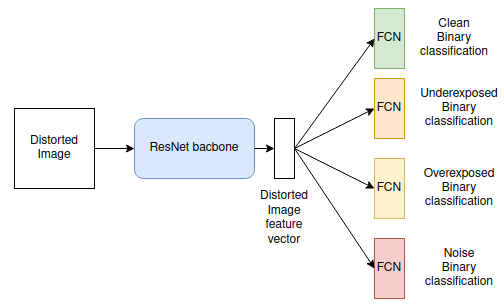}}
\caption{Framework to identify image distortion.}
\label{distortionidentification}
\end{figure}

\begin{figure*}[t!]
\includegraphics[width=\textwidth,height=4cm]{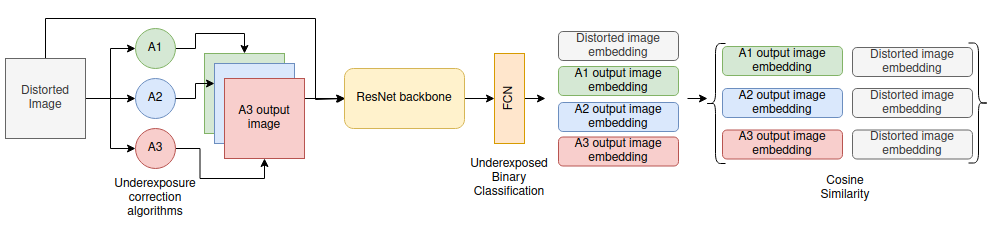}
\caption{An example of the framework to select the correction algorithm for distortion rectification. Assuming that in the first stage if the identified distortion is an ``underexposed image''.}
\label{algorithmselection}
\end{figure*}

\subsection{Architecture Details}
\label{Architecture Details}
Our algorithm deploys a two step procedure to first identify the distortion in the image and second to select an algorithm for rectifying the distortion Fig. \ref{distortionidentification} \& Fig. \ref{algorithmselection}. In the first step, the algorithm is responsible for identifying the distortion present in the image. It is an extension of the previous work \cite{Ameur2021} that makes use of the Multi-Task Learning (MTL) framework. Instead of using a DenseNet-169 as the backbone, we used ResNet-50 as the backbone network which is pretrained on ImageNet dataset. The features extracted from the distorted images are further sent to each head which are Fully Connected Networks. Each head is responsible for the binary classification of a single distortion type present in the image. As discussed earlier, such an architectural setting is capable to identify distortion types since the feature vector extracted from the distorted image captures information about the distortions present in the image.

Our model takes an input image and identifies the latest distortion that corrupted the image. For example, if an image has undergone two distortions wherein its exposure was compromised followed by noise being added, our algorithm would identify noise as the distortion type in the first pass and suggests an algorithm to denoise the image. Images with multiple distortions have to be passed into our model until it predicts the image to be clean. This problem closely resembles the process of sequential decision making.

In the second step, the framework is used to identify the subsequent correction algorithm that is necessary to rectify the distortion in the image. We perform a guided algorithm search based on the identified distortion thus restricting the search space to a limited number of algorithms.

Once the distortion type is identified, the search is performed only on those correction algorithms that address the identified distortion. The distorted image is passed through each of the relevant correction algorithms, the output of which is an image (Fig. \ref{algorithmselection}). Now each of these generated images are further passed through the ResNet-50 feature extractor and the corresponding head of the identified distortion, trained in first step, to generate corresponding algorithm feature vectors. These algorithm specific feature vectors have knowledge about the the presence or absence of the identified distortion and can be used as a good proxy to identify the correct candidate. For example, in Fig.\ref{algorithmselection}, if the identified distortion is an ``underexposed image'', then the feature vectors generated from the output of ``underexposure correction'' in the second stage can be processed to judge which algorithm has rectified the distortion (since each algorithm can have different degrees of correction). As a result, the distorted image feature vector has information about the distortions present in the original distorted image and the feature vector of the image generated from the correction algorithm has information about just the uncorrected distortions present in the original distorted image. In order to identify the correction algorithm, we perform a cosine similarity measure between each of the algorithm generated image feature vectors and the distorted image feature vector. The feature vector of the correction algorithm, that can correct the distorted image the most, should have a smaller cosine similarity with the feature vector of the distorted image since the angle between them should be relatively larger. Thus, we can essentially rank the correction algorithms according to their performance on the current input image using any similarity metric.

This entire procedure is repeated until our framework predicts the image to be clean.

In the next subsection, we describe the different algorithms used in our framework and discuss the training procedure in detail.

\subsection{Denoising Algorithms}
\label{Denoising Algorithms}
The aim of denoising algorithms is to filter out the additive noise from the distorted image and restore back the clean image. We use 3 variants of SCUNet as denoising algorithms. In recent years, the field of image restoration has seen a rise in the use of Transformer-based techniques. For combining the local modelling capability of residual convolutions with the non-local modelling capability of effective shifted window attention, SwinConv-Unet(SCUNet)~\cite{zhang2022practical}. This allows each pixel to contain global correlation while compensating for the drawbacks of conventional CNNs that only consider local receptive fields and ignore global features. Each variant of SCUNet is better suited for a particular noise range. We used SCUNet-15, SCUNet-25 and SCUNet-50 that were specialized to correct noise levels of 15 pixels, 25 pixels and 50 pixels in the image. 

\subsection{Exposure Correction Algorithms}
\label{Exposure Correction Algorithms}
We have used gamma correction algorithm with multiple gamma values as candidate exposure correction algorithms. Gamma correction is a non linear pixelwise transformation of input image. The intensity of the image pixels must first be scaled from [0, 255] to [0, 1]. After normalization, the power law equation as given by Eq.~\ref{eq:gamma_correction} is applied to get the output gamma corrected image, where $I_{in}$ will be input distorted image and $I_{out}$ will be output image. Afterwards, the output image is scaled back to the [0, 255] range. Gamma values less than 1 will make the image appear lighter (overexposed), while the image becomes darker when gamma values are greater than 1 (underexposed). The input image will not be affected by a gamma value of 1. In our experiments, we have used gamma values as inverse of the gamma values that are used to create the distortion. In theory, the inverse of the gamma values should restore the image back to its original format but due to clipping (in case when pixel values go beyond the true ranges) and other restraints we tend to see a drop in performance. We used the values of 0.33, 0.5, 1.25 and 5.0 for exposure correction.

\subsection{Training}
\label{Training}
We trained this entire approach in an end-to-end fashion. The ResNet-50 feature extractor was initialized with pretrained ImageNet weights and the Fully Connected Network (FCN) heads corresponding to different distortion types were randomly initialized (see Fig. \ref{distortionidentification}). We used the proposed dataset to train this setup to identify the distortion types present in the image. This enabled the ResNet-50 feature extractor and the FCN heads to learn to capture relevant information about the distortions present in the image. The input to the network was the distorted image and the output was the identification of the latest distortion that the image had undergone. We used Binary Cross Entropy Loss (BCE) for each of the FCN heads with label 1 for the latest distortion and 0 for the rest. We add the BCE losses for each head with equal weightage due to reasons stated in \cite{Ameur2021}. The FCN hidden units for each of the different distortion task was (2048, 512, 512, 1) with Leaky ReLU activation function. The learning rate was fixed to be 0.0001 for the entire training period using Adam optimizer with a weight decay of 0.0005. 

To summarise the training process with an example, if an image had undergone two levels of distortion namely, underexposure and addition of low-level noise, the network would at first predict low-level noise in the image. As a result the head corresponding to low-level noise would be regressed with the label=1 and the rest of the heads with label=0 and the sum of the losses would be used while backpropogating.

\section{Experiments}
\label{Experiments}
In this section, we first introduce four baselines and compare their performance with our approach on object detection as the downstream vision task. We use Pytorch's implementation of FasterRCNN \cite{FasterRCNN} as the object detector which reports a map score of \textbf{0.3959} on the clean images of our training dataset and \textbf{0.3955} on the clean images of our test dataset. Finally, we conduct a series of ablation experiments to analyse the behavior of the proposed framework with new distorted image datasets that the network had not been trained on. While comparing our approach with the four baselines, we make sure that the trainable parameters (including seed) and the training conditions remain similar amongst all.

\subsection{Hard-coded Distortion Classifier Model (HC-C)}
This baselines is inspired from the MAPF-Algorithms that train a image classification neural network to directly predict the choice of the algorithm. In this baseline we train a vanilla ResNet-50 feature extractor with a fully connected prediction network to classify the latest distortions present in the image. The ResNet-50 backbone was pretrained on ImageNet dataset and the last classification layer was initialized randomly. We use Cross Entropy Loss with a learning rate of 0.0001 alongside Adam optimizer with a weight decay of 0.0005. The hidden units of the prediction network comprise of (2048, 1024, 256, 64, 5) neurons with Leaky ReLU activation. The last layer comprises of 5 neurons because we consider 5 distortion types namely underexposed image, overexposed image, low-level noise image, high-level noise image and clean image. In case of algorithm selection, we mapped each of the distortions to the best correction algorithm for the identified distortion in the image. In a way, this is a rule based selection so we call this baseline Hard-Coded Distortion Classifier Model. 

\subsection{Random Multi-Task Learning Distortion Model (R-MTL)}
In this baseline, we extend the prior work \cite{Ameur2021} to train a vanilla ResNet-50 backbone pretrained on ImageNet dataset as a feature extractor. Subsequently, we train each head to identify a distortion type present in the image. Further, we map each of these heads to a set of candidate algorithms that can correct the distorted image. While choosing the correction algorithm, we uniformly sample one of the candidate correction algorithms (capable to correct the identified distortion type) to rectify the distorted image. The network architecture is the same as ~Fig.\ref{distortionidentification}. We trained 5 heads with hidden units comprising of (2048, 512, 512, 1) neurons with Leaky ReLU activation.

\subsection{Hard-Coded Multi-Task Learning Algorithm Selection Model (HC-MTL)}
This baseline's architecture is similar to the above Random Multi-Task Learning Distortion Model, however, based on the identified distortion type, we use the right algorithm for correcting the distortion. As a result, we treat this baseline as the \textbf{oracle} algorithm. Since this is a heuristic based algorithm selector, we name this baseline Hard-Coded Multi-Task Learning Algorithm Selection Model. 

\subsection{Fixed Pipeline}
We introduce another baseline for comparison wherein we fix the image processing sequence irrespective of the image type to replicate the most commonly used methodoly to do image processing for images in the dataset. The Fixed Pipeline 1 in our case is Exposure Correction (Gamma Correction with $\gamma=2.0$) followed by Denoising (SCUNet-15) and Fixed Pipeline 2 is the opposite arrangement of Fixed Pipeline 1. As discussed earlier, certain algorithm arrangements might not be suitable for a given goal task and can have deteriorating effects on performance as highlighted in case of Fixed Pipeline 2 (refer Table \ref{tab:Object Detection Performance}).

\subsection{Distortion Identification Accuracy}

The distortion identification accuracy of each of HC-C and MTL based methods is 94\% and 99.6\% on the train dataset respectively, and 84.99\% and 99.5\% on the test dataset respectively.

\begin{table*}[t!]
\caption{Comparison between mAP score and normalized mAP score of the baseline algorithms and our approach when Faster-RCNN is used as the object detection model. The normalized mAP score denotes the placement of the given approach on a normalised scale between Fixed Pipeline 1 (a lower bound) and the Oracle (the upper bound with prior knowledge of the distortion).}
\label{tab:Object Detection Performance}
\centering
\resizebox{\textwidth}{!}{%
\begin{tabular}{|l|l|l|l|l|}
\hline
\textbf{Algorithm} & \textbf{mAP score (Train)} & \textbf{mAP score (Test)} & \textbf{Normalised mAP (Train)} & \textbf{Normalised mAP (Test)}  \\ \hline
\begin{tabular}[c]{@{}l@{}}HC-MTL (Oracle) \end{tabular} & \textit{0.2444} & \textit{0.2432} & 100\% & 100\% \\ \hline
DeepClean & \textbf{0.2402} & \textbf{0.2395} & \textbf{95.1\%} & \textbf{96.2\%} \\ \hline
\begin{tabular}[c]{@{}l@{}}R-MTL\end{tabular} & 0.2238 & 0.2132 & 75.8\% & 69.2\% \\ \hline
\begin{tabular}[c]{@{}l@{}}HC-C\end{tabular} & 0.2234 & 0.2001 & 75.3\% & 55.8\% \\ \hline
\begin{tabular}[c]{@{}l@{}}Fixed Pipeline 1\end{tabular} & 0.1593 & 0.1457 & 0\% & 0\% \\ \hline
\begin{tabular}[c]{@{}l@{}}Fixed Pipeline 2\end{tabular} & 0.0 & 0.0 & - & - \\ \hline
\end{tabular}%
}
\end{table*}

\subsection{DeepClean and Baselines Discussion}
From Table~\ref{tab:Object Detection Performance}, it is apparent that our proposed algorithm, DeepClean outperforms other baselines by a significant margin on the train and test dataset. 
HC-MTL, R-MTL and DeepClean are fairly resilient to newer forms of distortions of the same kind (see discussion above). It is also worth noting that DeepClean framework is not trained on algorithm generated images however it is still able to identify a good correction algorithm to restore the image back. Since, DeepClean does not impose any restrictions or assumptions on the set of candidate algorithms it becomes possible to introduce new algorithms directly into the framework which makes it a plug-and-play module.

\section{Conclusion \& Future Work}
In this paper, we have proposed a framework for distortion identification and selecting a corresponding correction algorithm to rectify the identified distortion. The proposed framework comprises of a feature extractor that is shared across the distortion identification network. The feature extractor learns to extract important information that is relevant to the different distortions present in the image. Each head in the identification network makes use of these captured features to identify a unique type of distortion in the input image. 
The experimental results showed that our proposed
framework provides better performance on the object detection task. The choice of framework not only allows it to be scalable with respect to the number of distortion types but also with their corresponding candidate algorithms. As future work, firstly we are investigating ways to learn to capture better feature representations learnt by the network that can further aid the framework. Secondly, we intend to extend it to other kinds of distortions and correction algorithms. Finally, we want to investigate the use of autoregressive architectures like LSTMs \cite{lstm} and Transformers \cite{transformer} for distortion identification and algorithm selection



\bibliography{mybibfile}

\end{document}